# CUDA optimized Neural Network predicts blood glucose control from quantified joint mobility and anthropometrics

Sterling Ramroach
Department of Electrical and
Computer Engineering
The University of the West Indies, St.
Augustine campus.
8684940866
sramroach@gmail.com

Andrew Dhanoo
Department of Life Sciences
The University of the West Indies, St.
Augustine campus.
868-662-2002 ext 83090
andrewdhanoo@hotmail.com

Brian Cockburn
Department of Life Sciences
The University of the West Indies, St.
Augustine campus.
868-662-2002 ext 83541
brian.cockburn@sta.uwi.edu

Ajay Joshi
Department of Electrical and
Computer Engineering
The University of the West Indies, St.
Augustine campus.
868-662-2002 ext 83144
ajay.joshi@sta.uwi.edu

## ABSTRACT

Neural network training entails heavy computation with obvious bottlenecks. The Compute Unified Device Architecture (CUDA) programming model allows us to accelerate computation by passing the processing workload from the CPU to the graphics processing unit (GPU). In this paper, we leveraged the power of Nvidia GPUs to parallelize all of the computation involved in training, to accelerate a backpropagation feed-forward neural network with one hidden layer using CUDA and C++. This optimized neural network was tasked with predicting the level of glycated hemoglobin (HbA1c) from non-invasive markers. The rate of increase in the prevalence of Diabetes Mellitus has resulted in an urgent need for early detection and accurate diagnosis. However, due to the invasiveness and limitations of conventional tests, alternate means are being considered. Limited Joint Mobility (LJM) has been reported as an indicator for poor glycemic control. LJM of the fingers is quantified and its link to HbA1c is investigated along with other potential non-invasive markers of HbA1c. We collected readings of 33 potential markers from 120 participants at a clinic in south Trinidad. Our neural network achieved 95.65% accuracy on the training and 86.67% accuracy on the testing set for male participants and 97.73% and 66.67% accuracy on the training and testing sets for female participants. Using 960 CUDA cores from a Nvidia GeForce GTX 660, our parallelized neural network was trained 50 times faster on both subsets, than its corresponding CPU implementation on an Intel® Core™ i7-3630QM 2.40 GHz CPU.

## CCS Concepts

• **Applied computing ~ Health informatics** • **Applied computing ~ Bioinformatics** • Computing methodologies ~ Massively parallel algorithms • Computing methodologies ~ Artificial intelligence.

## Keywords

Neural Network; CUDA; Parallel Processing; Limited Joint Mobility; Diabetes Mellitus.

## 1. INTRODUCTION

An Artificial Neural Network (NN) is a model that was inspired by the biological neural network from the brains of animals [1]. The NN is made up of various layers (input, hidden, and output) of neurons where each neuron performs a calculation using inputs from the data source or the outputs of earlier neurons. Neurons are connected by weighted edges. These weights adjust as learning progresses. After the NN is exposed to training data, it can be used to solve many challenging problems. However, NNs require a large amount of computation in order to adequately adjust the weights in the network. This problem is exacerbated by the inclusion of hidden layers in the network. The number of hidden layers needed is dependent on the data used to train the network. If the data is linearly separable then there may not be a need for any hidden layers. This can occur for binary classification problems or regression problems with a high correlation between a subset of attributes and the target. When there is a continuous mapping between finite spaces and a separator more complex than linear is required, one hidden layer may be used. Although quite rare, the usage of 2 or more hidden layers allows a NN to represent any function of any shape. The computation needed for the network to learn or adjust its weights grow exponentially as more layers are added. The NN architecture is made even more complex when the number of neurons which comprise each layer is chosen. If too few neurons are used in the hidden layers, the NN will suffer from underfitting. This will lead to an inability to adequately observe the signals in the dataset. On the other end of the spectrum, the usage of too many neurons will result in overfitting. Apart from a significant increase in processing time, the weights in the NN will be too tightly coupled to the training data. The NN will not be able to properly generalize for cases not present in the training data. Training a NN is a computationally intense problem.

The Graphics Processing Units (GPU) is a highly parallel, many-core processor with high memory bandwidth and compute capability [2]. GPUs were used for graphics applications before high-performance computing researchers realized that it can also be used for simple repetitive calculations [3]. The Compute Unified Device Architecture (CUDA) is a parallel programming model developed by the NVIDIA Corporation to facilitate high performance computing on any system with a CUDA-enabled GPU. This type of functionality is commonly known as general-purpose programming on a GPU (GPGPU) [4]. The difference between design philosophies of GPUs and CPUs are illustrated in Figure 1. GPUs are core to advancing deep learning, as deep architectures are perfectly suited to parallelism. There are two types of parallelism: data-based [5] and model-based [6]. Data-based parallelism distributes large datasets over different nodes. Model-based parallelism splits complex models across different nodes. Learning in these networks must account for coordination and communication among processors.

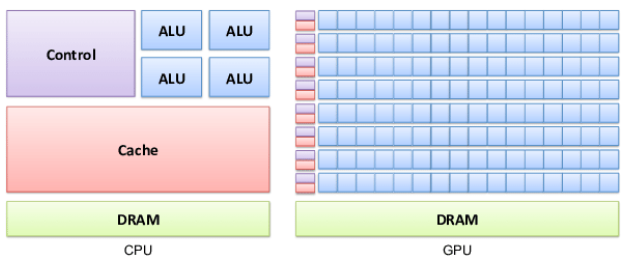

**Figure 1. CPU vs GPU architecture and design philosophy [7].**

The CUDA programming model facilitates the development of software that transparently scales its parallelism to leverage the increasing number of processor cores [2]. This allows the same program to run on any CUDA GPU. When applied to neural networks, the computation must be partitioned into coarse sub-calculations that can be solved independently and in parallel by blocks of threads on any of the available multiprocessors within a GPU [8]. This is facilitated through the use of the CUDA C++ programming language. CUDA C++ is an extension of the standard C++ language. New keywords are added to allow data parallel functions and communication with the processor cores in the GPU. The data parallel functions in CUDA are called kernels. Kernel functions exploit data parallelism by generating a large number of threads. Each thread executes code on a separate element. This is applicable to the training of neural networks as it is a problem which exhibits data parallelism.

Bioinformatics and the health sector are popular avenues where neural networks are applied. Artificial intelligence and machine learning has been leveraged for a variety of classification tasks including predicting the risk of T2DM from electronic medical records [9]. We are now using these tools to predict the level of HbA1c from non-invasive markers.

The global estimate of the number of adults living with diabetes is 246 million with projections of up to 380 million by the year 2025 [9]. Every year, up to 10% of pre-diabetic patients develop Type 2 (T2) diabetes mellitus (DM) [10], therefore early detection and treatment is necessary to reduce the progression of the disease. DM may be diagnosed using an HbA1c (glycated hemoglobin) test, which when levels are greater than 6.5% (48 mmol/mol) [11] a diagnosis of poor glucose control may be given. HbA1c testing may be expensive or unavailable, especially for low income patients in developing countries. HbA1c has also been reported to not adequately discriminate between Type 1 (T1) DM and healthy populations in the absence of overt hyperglycemia [12] and correspondingly cannot be used with enough accuracy for the early detection of diabetes complications [13]. HbA1c as well as most other common methods of diabetes testing requires a blood sample to be collected, which may present a barrier to some patients as the willingness to seek testing or adherence to regular monitoring may be impaired [14], particularly for children [15]. These factors limit the use of HbA1c as a screening tool for diabetes in large populations, therefore, reliable and convenient non-invasive methods are needed to perform first line screening to identify persons with prediabetes or diabetes. Limited Joint Mobility syndrome (LJM) is a long term consequence of DM [16]. LJM is diagnosed by the progression of the stiffness in joints of the hands and fingers, impaired grip strength, and fixed flexion contractures of the smaller hand joints. It has been shown that patients with diabetes are more likely to develop musculoskeletal disorders such as LJM, and carpal tunnel syndrome [17].

According to Rosenbloom and Frias [16] and Benedetti et al. [18], fixed flexion contractures are observed in children with T1DM. LJM can develop in up to 26% of patients without diabetes but for patients with DM, up to 58% show signs of LJM [19]. Pressing the palms of both hands against each other with maximally flexed wrists is part of a clinical test which strongly supports the diagnosis of LJM [20]. If it is impossible for the surface of all fingers on one hand to be in contact with the surface of all fingers of the other hand, as measured by the joints making contact, then this indicates the existence of flexion contractures of the fingers, which itself is an indication of LJM. Recording these readings result in large amounts of data. The use of machine learning (ML) can aid in data analytics to provide interesting insights.

Given the significant relationship between DM and LJM, the primary aim of this study is to find supporting evidence for a non-invasive test using LJM readings to estimate blood glucose control. All angles are used in conjunction with other non-invasive measurements to train a neural network on the CPU and GPU. The male and female datasets contained readings for 61 and 59 participants, respectively. The training data comprise 75% of the dataset and the remaining 25% of the data was used to test the neural network. The secondary aim of this study is to highlight the extent to which CUDA can be used to optimize neural networks.

## 2. Methods
### 2.1 Data Collection
Data and samples were collected from 120 participants at diabetes awareness drives at a public primary care health center in Trinidad and Tobago. All attendees were included in the study. Ethics approval was granted by the campus ethics committee of the University of the West Indies, Faculty of Medical Sciences, St. Augustine Campus. Participants gave written consent to have samples collected, analyzed and reported.

Measurements of the angles made by each joint of each finger of the dominant hand, when pressed against the palmar surface of the other hand, with wrists maximally flexed were recorded using a finger goniometer (baseline). Capillary blood samples were taken to measure HbA1c using a point of care HbA1c Analyzer (SD Biosensor). Patient medical history, demographics (date of birth, gender and ethnicity) and anthropometrics (height, weight, circumferences of the waist, hip, neck, wrists and ankles) were measured. Additional features were created such as body mass index (BMI), waist to hip ratio (WHRatio), wrist to waist ratio (right wrist = WRWRatio and left wrist = WLWRatio), wrist to hip ratio (WRHRatio and WLHRatio) and a binary column indicating whether or not the participant was on diabetes medication (onMed) which included oral hypoglycemics and/or insulin. All recordings culminated in a dataset of 120 rows and 33 columns. This data is provided in Supplementary Table 1. The full dataset was split into male and female subsets since sexual dimorphism was demonstrated as variable for the occurrence of T2DM [21].

Each finger is represented by X1 (first or little finger), X2 (second), X3 (third), X4 (fourth), and X5 (fifth or thumb finger). The joints were denoted by MCP (metacarpophalangeal joint), PIP (proximal interphalangeal), and DIP (distal interphalangeal). Only the MCP and IP (interphalangeal) joints were considered for X5. Measurements of the angles were taken in relation to a normal reading of 0 degrees (straight fingers). A negative number represents hyper-extension (when the joint bends backwards) and a positive number indicates the angle each joint makes if it does not straighten (when the joint bends forwards or into the palm).

### 2.2 CUDA-optimized Neural Network
The network used to simulate backpropagation is the multi-layer perceptron feed forward NN. Each neuron of each layer is connected to each neuron of adjacent layers. Supervised learning is achieved with the use of gradient descent. Gradient descent is performed via backpropagation to find combinations of connection weights between the layers to map input values to output values. In the feed-forward network, each neuron in a layer takes its input from the outputs of every neuron in an earlier layer. The input layer is fed into the network from the data source. A subsequent layer such as the first hidden layer, takes its input from the outputs of all neurons in the input layer. Each neuron uses an activation function to convert the input signal to an output signal. The traditional activation functions are the logistic sigmoid and hyperbolic tangent. The sigmoid is an S-shaped curve within the range 0 and 1 and is displayed in Equation 1.

$$sigmoid(x) = \frac{1}{1 + e^{-x}}$$

**Equation 1. Sigmoid function.**

This function is non-linear in nature, which implies that combinations of it are non-linear. This makes it a good candidate function when layers need to be stacked. Due to the steepness of its plot, small changes in the input can cause significant changes to the output. This property makes the sigmoid suitable for classification. The drawbacks of this function are the vanishing gradient problem (when large regions of input are mapped to a small range, the gradient becomes too small to result in change), the saturation of gradients, it is not zero centered, and its slow convergence. However, given the task and the data, the sigmoid is adequate for a successful NN. The learning rate of the NN was set to 0.1 with no momentum.

The parallelization approach taken is a topological data parallel approach. CUDA is used to deploy the neurons to the GPU. There is only one copy of the NN which exists on the GPU. Each neuron is independently executed on its own thread. The input data is stored in a one dimensional float array where the length of the array is equal to the number of instances multiplied by the number of attributes in the dataset (i.e., length = rows × columns). The weights for each layer are also stored in one dimensional float arrays. For example, the size of the weight array connecting the input layer to the hidden layer is equal to the number of neurons in hiddenlayer1 × columns × sizeof(float).

After each feed forward iteration of the network, the difference between the output values and the true values are used for the backpropagation step. Figure 2 illustrates the flow of data for training and testing the NN. After normalizing the data and other pre-processing tasks, the data is split into a training and a testing set. The training set is comprised of 75% of the data and the testing set is made up of the remaining 25%. After allowing the NN to be trained for a certain number of epochs, it is tasked with classifying the instances of the testing set.

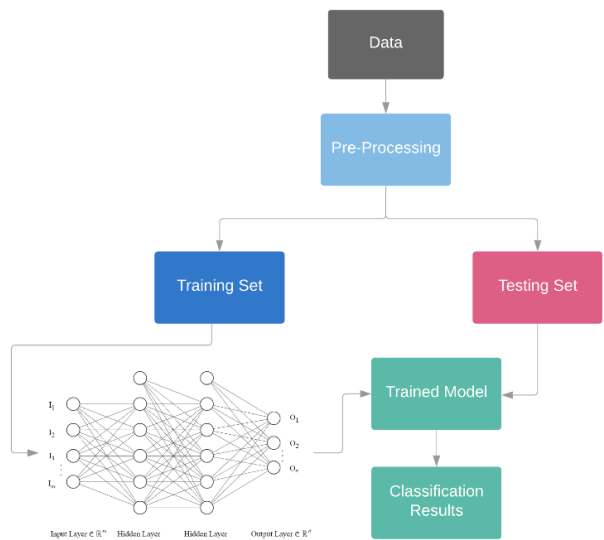

**Figure 2. Flow of data**

## 3. Results and Discussion

The relationship between musculoskeletal complications and HbA1c is still being explored. LJM may act as an independent marker for HbA1c levels. The investigation of this hypothesis is hindered by many factors, including: inaccurate reporting of time since diagnosis due to late detection, and the size of the dataset. Some studies showed no link between hyperglycemia and LJM of the hands [22], whereas other studies have found significant correlations [23]. LJM was observed in 86.6% of participants as compared to ranges of up to 76% reported in previous works [23].

The primary aim of this study is to find supporting evidence for a non-invasive test using LJM to estimate blood glucose control. Figures 3 and 4 illustrate the accuracy of the NN for both the training and testing sets as the number of epochs vary.

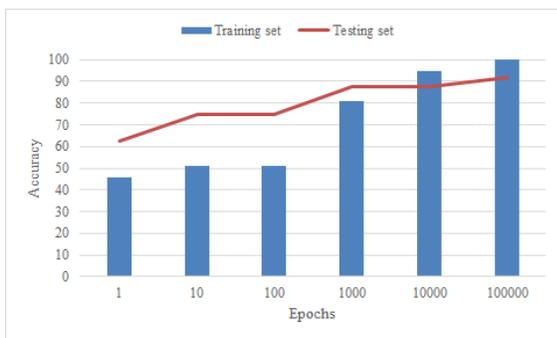

**Figure 3. Accuracy of the neural network predictions on the male subset.**

At 100,000 epochs on the male subset, the NN was able to achieve a 100% accuracy at classifying instances in the training set. This also led to an accuracy of 91.7% on the testing set. After a similar number of epochs, 97.7% of the female training set was correctly classified, however, only 66.6% of the female test set was accurately classified.

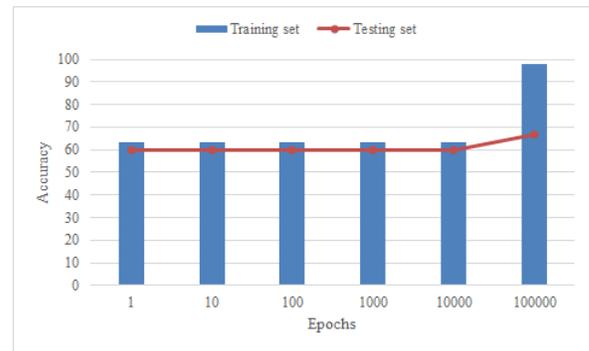

**Figure 4. Accuracy of the neural network predictions on the female subset.**

There may be stronger correlations between the attributes in the dataset and the HbA1c level of the male participants, as opposed to the female participants. The NN was clearly able to generalize for the instances in the male testing set, but not the female testing set. The reason for a high accuracy on the female training set at 100,000 epochs is undone by a low accuracy on the testing set. This is due to the NN being coupled too tightly to the training set, also known as overfitting the training data. These preliminary results suggest that LJM can be considered when screening male participants only. Future work will include collecting samples from a larger number of participants so that more conclusive tests can be performed. The primary aim of this work is therefore achieved for males, but no supporting evidence is found when investigating females.

The secondary aim of this work is achieved by comparing the time taken to train the same NN on the CPU versus the GPU. Figures 5 and 6 illustrate the vast difference in training time for both the male and female subsets.

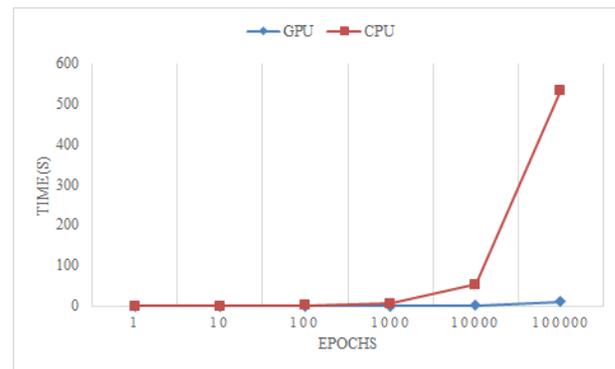

**Figure 5. Time to train neural network on the male subset with and without CUDA.**

The CUDA-optimized NN was trained 50 times faster on both male and female subsets than the sequential NN. As the dataset grows in size, this difference in performance would exponentially increase.

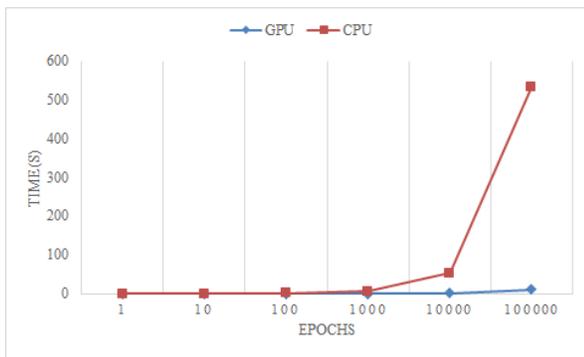

**Figure 6. Time to train neural network on the female subset with and without CUDA.**

## 4. Conclusion

These results indicate that a differentiation of good from poor glucose control can be made on males via non-invasive readings which include limited joint mobility readings. Our neural network was unable to successfully classify the female subset of participants. This can be due to the size of the dataset. Further research is needed to determine the feasibility of applying these findings to the wider population as the use of a larger sample size would provide more conclusive results. The Compute Unified Device Architecture framework was used to optimize the neural network training time. Computation was passed from the CPU to the GPU during training of the neural network. This resulted in a speedup of 50 times that of the sequential neural network.

## 5. ACKNOWLEDGMENTS

Our thanks to the Diabetes Association of Trinidad and Tobago for organizing the clinics for data collection. Special thanks to Mr. Craig Subnarine for his scholarly advice.

## 6. REFERENCES

[1] Jain, A. K., Mao, J. and Mohiuddin, K. M. Artificial neural networks: A tutorial. Computer, 29, 3 (1996), 31-44.

[2] Sooknanan, D. J. and Joshi, A. GPU computing using CUDA in the deployment of smart grids. IEEE, City, 2016.

[3] Mei, X. and Chu, X. Dissecting GPU memory hierarchy through microbenchmarking. IEEE Transactions on Parallel and Distributed Systems, 28, 1 (2017), 72-86.

[4] Sanders, J. and Kandrot, E. CUDA by example: an introduction to general-purpose GPU programming. Addison-Wesley Professional, 2010.

[5] Heywood, P., Maddock, S., Casas, J., Garcia, D., Brackstone, M. and Richmond, P. Data-parallel agent-based microscopic road network simulation using graphics processing units. Simulation Modelling Practice and Theory, 83 (2018), 188-200.

[6] Divya, U. V. and Prasad, P. S. Hashing Supported Iterative MapReduce Based Scalable SBE Reduct Computation. Springer, City, 2018.

[7] Lounis, M., Bounceur, A., Laga, A. and Pottier, B. GPU-based parallel computing of energy consumption in wireless sensor networks. IEEE, City, 2015.

[8] Nvidia, C. Nvidia cuda c programming guide. Nvidia Corporation, 120, 18 (2011), 8.

[9] Mani, S., Chen, Y., Elasy, T., Clayton, W. and Denny, J. Type 2 diabetes risk forecasting from EMR data using machine learning. American Medical Informatics Association, City, 2012.

[10] Inzucchi, S. E. and Sherwin, R. S. The prevention of type 2 diabetes mellitus. Endocrinol Metab Clin North Am, 34, 1 (Mar 2005), 199-219, viii.

[11] American Diabetes, A. Standards of Medical Care in Diabetes-2018 Abridged for Primary Care Providers. Clin Diabetes, 36, 1 (Jan 2018), 14-37.

[12] Cordelli, E., Maulucci, G., De Spirito, M., Rizzi, A., Pitocco, D. and Soda, P. A decision support system for type 1 diabetes mellitus diagnostics based on dual channel analysis of red blood cell membrane fluidity. Computer methods and programs in biomedicine, 162 (2018), 263-271.

[13] Florkowski, C. HbA1c as a diagnostic test for diabetes mellitus–reviewing the evidence. The Clinical Biochemist Reviews, 34, 2 (2013), 75.

[14] Wright, S., Yelland, M., Heathcote, K., Ng, S.-K. and Wright, G. Fear of needles-nature and prevalence in general practice. Australian family physician, 38, 3 (2009), 172.

[15] Cemeroglu, A., Can, A., Davis, A., Cemeroglu, O., Kleis, L., Daniel, M., Bustraan, J. and Koehler, T. Fear of needles in children with type 1 diabetes mellitus on multiple daily injections and continuous subcutaneous insulin infusion. Endocrine Practice, 21, 1 (2014), 46-53.

[16] Rosenbloom, A. L., Grgic, A. and Frias, J. L. Diabetes mellitus, short stature and joint stiffness—a new syndrome. Pediatric Research, 8, 4 (1974), 441.

[17] Pandey, A., Usman, K., Reddy, H., Gutch, M., Jain, N. and Qidwai, S. Prevalence of hand disorders in type 2 diabetes mellitus and its correlation with microvascular complications. Annals of medical and health sciences research, 3, 3 (2013), 349-354.

[18] Benedetti, A. and Noacco, C. Juvenile diabetic cheiroarthropathy. Acta diabetologia latina, 13, 1-2 (1976), 54-67.

[19] Gerrits, E. G., Landman, G. W., Nijenhuis-Rosien, L. and Bilo, H. J. Limited joint mobility syndrome in diabetes mellitus: A minireview. World journal of diabetes, 6, 9 (2015), 1108.

[20] Sauseng, S., Kästenbauer, T. and Irsigler, K. Limited joint mobility in selected hand and foot joints in patients with type 1 diabetes mellitus: a methodology comparison. Diabetes, nutrition & metabolism, 15, 1 (2002), 1-6.

[21] Gale, E. A. and Gillespie, K. M. Diabetes and gender. Diabetologia, 44, 1 (2001), 3-15.

[22] Aydeniz, A., Gursoy, S. and Guney, E. Which musculoskeletal complications are most frequently seen in type 2 diabetes mellitus? Journal of International Medical Research, 36, 3 (2008), 505-511.

[23] Mustafa, K. N., Khader, Y. S., Bsoul, A. K. and Ajlouni, K. Musculoskeletal disorders of the hand in type 2 diabetes mellitus: prevalence and its associated factors. International journal of rheumatic diseases, 19, 7 (2016), 730-735